\documentclass[10pt]{article} 

\usepackage[preprint]{rlj} 

\usepackage{amssymb}            
\usepackage{mathtools}          
\usepackage{mathrsfs}           
\usepackage{graphicx}           
\usepackage{subcaption}         
\usepackage[space]{grffile}     
\usepackage{url}                
\usepackage{lipsum}             

\usepackage{tabularx} 
\usepackage{array, booktabs} 

\usepackage[vlined,linesnumbered,ruled]{algorithm2e}        
\usepackage{wrapfig}
\usepackage{multirow}

\usepackage{amsmath,amsfonts,bm, bbm}

\def\Figref#1{Figure~\ref{#1}}

\def\Secref#1{Section~\ref{#1}}
\def\Appref#1{Appendix~\ref{#1}}

\def\eqref#1{equation~\ref{#1}}

\def\Algref#1{Algorithm~\ref{#1}}

\def\Tabref#1{Table~\ref{#1}}

\def\1{\bm{1}}

\def\vh{{\bm{h}}}

\def\vo{{\bm{o}}}

\def\vr{{\bm{r}}}
\def\vs{{\bm{s}}}

\def\mQ{{\bm{Q}}}

\def\mV{{\bm{V}}}

\DeclareMathAlphabet{\mathsfit}{\encodingdefault}{\sfdefault}{m}{sl}
\SetMathAlphabet{\mathsfit}{bold}{\encodingdefault}{\sfdefault}{bx}{n}

\newcommand{\E}{\mathbb{E}}

\newcommand{\R}{\mathbb{R}}

\newcommand{\pref}{\bm{\omega}}

\DeclareMathOperator*{\argmax}{arg\,max}

\newcommand{\blue}[1]{\textcolor{blue}{#1}}
\newcommand{\camera}[1]{#1}

\newcommand\blfootnote[1]{%
  \begingroup
  \renewcommand\thefootnote{}\footnote{#1}%
  \addtocounter{footnote}{-1}%
  \endgroup
}

\newcommand{\tablesize}{\scriptsize} 
\newcommand{\cisize}{\fontsize{6.5pt}{7.5pt}\selectfont} 

\title{Momba: Network Modernization Improves Multi-Objective Reinforcement Learning}

\setrunningtitle{Momba: Network Modernization Improves Multi-Objective Reinforcement Learning}

\author{
Adam Štafa\textsuperscript{1}, Santeri Heiskanen\textsuperscript{2}, 
Petr Novotný\textsuperscript{1}, Joni Pajarinen\textsuperscript{2}
}

\emails{stafa@mail.muni.cz, petr.novotny@fi.muni.cz \\ \{santeri.heiskanen,joni.pajarinen\}@aalto.fi}

\affiliations{
$^{1}$\textbf{Faculty of Informatics, Masaryk University}\\
$^{2}$\textbf{Department of Electrical Engineering and Automation, Aalto University}\\
}

\contribution{
    We showcase that employing more expressive neural network architectures with a distributional critic in a MORL algorithm substantially 
    improves the performance \camera{on continuous control tasks without requiring complex preference selection algorithms or MORL-specific update rules.}
    }
    {
    \camera{Previous works in MORL achieved improvements in sample efficiency and asymptotic performance by employing advanced preference selection algorithms \citep{alegre_sample-efficient_2023} or specialized update rules \citep{yang_generalized_2019, Li_cola_2025}.}
    \camera{In single-objective RL}, in contrast, a range of studies \citep{nauman_bro_2024,lee_simba_2024,lee_hyperspherical_2025, palenicek_scaling_2025, palenicek_xqc_2026} has
    shown that the combination of (i) feature normalization, (ii) weight normalization, 
    and (iii) categorical critic loss can improve the sample efficiency and asymptotic 
    performance of existing reinforcement learning algorithms, without altering them. 
    We apply the same principles to multi-objective reinforcement learning and showcase 
    that improvements in the architecture outperform selected baselines.
    Additionally, we study these design choices and identify the distributional critic as a major component.
    } 
\contribution{
    We propose a simple adaptation of the categorical critic to the multi-objective domain by directly learning 
    to predict the scalarized return distribution, motivated by the commonplace scalarized expected return
    objective.
    }
    {
    Previous research applying the distributional critic \citep{bellemare_distributional_2017} 
    to multi-valued return functions by modeling the multivariate joint distribution of rewards 
    has been limited to tabular cases \citep{wiltzer_foundations_2024}, or required specifying
    a kernel function for measuring distance between distributions \citep{zhang_distributional_2021}.
    Instead, we specifically target multi-objective reinforcement learning, under the 
    prevalent linear scalarization assumption 
    \citep{abels_dynamic_2019, xu_prediction-guided_2020, kyriakis_pareto_2022, basaklar_pdmorl_2023, alegre_sample-efficient_2023}, and propose a simple approach where the Q-network directly predicts the scalarized
    return distribution. Through extensive empirical studies, we validate the effectiveness of the 
    proposed approach.
    }


\summary{
Recent advances in deep reinforcement learning (RL) have shown that improving neural network architectures can yield substantial gains in sample efficiency and asymptotic performance without altering the underlying algorithms. In contrast, work on multi-objective reinforcement learning (MORL), which aims to discover a set of policies that balance trade-offs among conflicting objectives, has predominantly focused on algorithmic innovations, leaving the area of architectures underexplored. While the optimal policies and value functions can differ significantly depending on the trade-offs, MORL algorithms commonly represent them with simple feedforward networks conditioned on the trade-off. This raises the question of whether the performance of the algorithms could be improved with more expressive function approximators. In this paper, we integrate recent advances in neural network design: (i) observation and feature normalization, (ii) weight normalization, and (iii) modeling of distributional returns with an entropy-regularized MORL algorithm. The empirical results across standard continuous control benchmarks demonstrate that these changes substantially improve the quality of the produced solution sets without requiring major changes to the underlying algorithm.

}

\begin{document}

\makeCover  
\maketitle  

\begin{abstract}
Recent advances in deep reinforcement learning (RL) have shown that improving neural network architectures can yield substantial gains in sample efficiency and asymptotic performance without altering the underlying algorithms. In contrast, work on multi-objective reinforcement learning (MORL), which aims to discover a set of policies that balance trade-offs among conflicting objectives, has predominantly focused on algorithmic innovations, leaving the area of architectures underexplored. While the optimal policies and value functions can differ significantly depending on the trade-offs, MORL algorithms commonly represent them with simple feedforward networks conditioned on the trade-off. This raises the question of whether the performance of the algorithms could be improved with more expressive function approximators. In this paper, we integrate recent advances in neural network design: (i) observation and feature normalization, (ii) weight normalization, and (iii) modeling of distributional returns with an entropy-regularized MORL algorithm. The empirical results across standard continuous control benchmarks demonstrate that these changes substantially improve the quality of the produced solution sets without requiring major changes to the underlying algorithm.

\end{abstract}

\begin{figure}[htb]
    \centering
    \includegraphics[alt={Normalized Hypervolume (left) and normalized EUM (center) of Momba, PGMORL, CAPQL and DPMORL, aggregated over all environments. Momba outperforms the runner-up PGMORL in both metrics by a decent margin. The rightmost plot showcases the generated solution sets in Ant, where Momba creates the solution set with best convergence and coverage.},width=\linewidth]{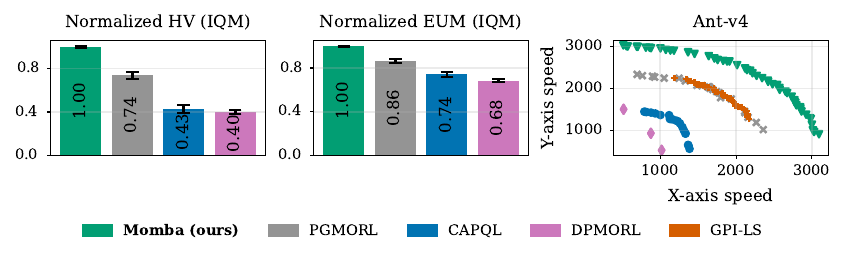}
    \caption{
        \textbf{Momba improves performance on 7 continuous control tasks}. 
        Left and center columns show the normalized hypervolume (HV) and expected
        utility (EUM) (IQM and 95\% SBCIs over 10 seeds), aggregated over 7 tasks.
        Momba outperforms both the runner-up, PGMORL, and CAPQL, our underlying
        algorithm.
        We report GPI results separately in \Figref{fig:gpi_comparison} due to a different evaluation protocol.
        The right column displays the generated solution sets in 
        the challenging Ant environment, demonstrating that Momba can generate solutions
        with good coverage.
    }
    \label{fig:main_fig}
\end{figure}
\section{Introduction}\label{sec:introduction}

Multi-Objective Reinforcement Learning (MORL) offers a principled approach
for identifying a set of policies, considering different trade-offs between
multiple, conflicting objectives \citep{roijers_survey_2013}, often 
seen in real-world problems such as drug design \citep{olivecrona_molecular_2017, zhou_optimization_2019}, 
medical treatment \citep{jalalimanesh_multi-objective_2017}, or even robotics \citep{haarnoja_composable_2018, xie_drl_robotic_2019}. 
While single-objective reinforcement learning (SORL) can achieve similar results
by encoding the trade-off into the reward function \citep{skalse_reward_gaming_2022, muslimani_reward_design_2025},
MORL allows one to make an informed decision on the desired trade-off \emph{after}
the policies are generated \citep{hayes_practical_2022}. Existing work 
has developed various learning algorithms for MORL \citep{abels_dynamic_2019, xu_prediction-guided_2020, lu_capql_2023, cai_distributional_2023},
yet many of them still suffer from poor sample efficiency \citep{Li_cola_2025}.
\blfootnote{Code to reproduce the experiments is available at \url{https://github.com/adamstafa/momba}}
\blfootnote{Correspondence to: \href{mailto:stafa@mail.muni.cz}{stafa@mail.muni.cz}}

One approach for improving sample efficiency in MORL is to learn a single policy,
conditioned on the desired trade-off \citep{yang_generalized_2019, basaklar_pdmorl_2023}.  
In theory, efficient information sharing via a single policy should speed up learning, yet
in practice, this approach faces significant challenges. As the behaviors between different 
trade-offs may differ significantly, the policy's ability to generalize is hindered. 
The previous work on MORL largely addressed this issue by developing new algorithmic frameworks while 
customarily using simple feedforward networks for the conditioned value function 
and policy representation \citep{yang_generalized_2019, lu_capql_2023}.
At the same time, recent advances in single-objective RL (SORL) have demonstrated
that enhanced neural network architectures can significantly improve performance and sample efficiency
\citep{nikishin_primacy_2022, nauman_bro_2024, lee_simba_2024, lee_hyperspherical_2025, palenicek_scaling_2025, palenicek_xqc_2026}. 
Therefore, we raise a natural question: 
\emph{Could multi-objective reinforcement learning benefit from recent advances in neural network design for deep RL?}

In this paper, we study this question by building on top of SimbaV2 \citep{lee_hyperspherical_2025},
a recent architecture designed for deep RL that utilizes (i) observation \& feature normalization,
(ii) weight normalization, and (iii) distributional critic to improve the training dynamics of neural networks.
To adapt the distributional critic to the multivariate return distribution, we propose to learn the scalarized distributional returns.
We apply the SimbaV2 architecture 
with the proposed changes on top of an existing entropy-regularized MORL algorithm, 
CAPQL \citep{lu_capql_2023}, and perform extensive evaluations in 7 
continuous control tasks, commonly used in MORL research. Our findings demonstrate that the
performance of an existing MORL algorithm can be significantly boosted by adopting the above-mentioned
techniques to the neural networks used to represent conditioned value functions
and policies. Moreover, we perform comprehensive ablations on our design choices and uncover 
that the distributional critic is a central component behind the performance improvement.

\section{Related Work}\label{sec:related_work}
Our work is focused on advancing MORL \camera{in continuous control tasks} by using techniques from scaling 
of deep RL and modeling of multivariate distributional returns. Thus, various parts of our work exist 
in the literature.

\textbf{Multi-objective reinforcement learning} can be roughly divided into three fields:
1) single-policy methods, 2) multi-policy methods, and 3) general policy methods. 
The single-policy methods consider only a single preference, and then transform the problem to a single objective problem, making it difficult to adapt to new, unseen preferences 
\citep{roijers_survey_2013, hayes_practical_2022, roijers_multi-objective_2018}. 
Multi-policy methods seek to find a set of solutions by training separate policies for different 
trade-offs, which often results in poor sample efficiency 
\citep{roijers_survey_2013,  hayes_practical_2022, xu_prediction-guided_2020}. 
General policy methods sidestep these issues by learning a universal policy, often 
conditioned on the trade-off, which is used to approximate the whole Pareto front 
\citep{abels_dynamic_2019, yang_generalized_2019, basaklar_pdmorl_2023}. 
Related to our work, \cite{shu_hypermorl_2024} use a trade-off conditioned hypernetwork 
for generating the weights for the policy, while \citet{Li_cola_2025} learn a common latent space using 
self-consistency loss to avoid redundant learning under different trade-offs. However, these methods
also introduce changes to algorithmic components, while we explicitly \camera{change only the critic loss 
and preference sampling frequency.}

\textbf{Specializing architectures for RL} is a recent trend that aims to boost the performance
and sample efficiency of existing RL algorithms by updating the neural network designs. 
The approaches have considered biasing the network to use simpler features for
predictions \citep{lee_simba_2024}, pairing strong regularization with 
optimistic exploration \citep{nauman_bro_2024}, or even designing a network
that improves the loss landscape of the Bellman error \citep{palenicek_xqc_2026}. Regardless, to our knowledge,
the application of these ideas to the multi-objective domain has been limited,
with the exception of GPI-LS \citep{alegre_sample-efficient_2023}, which used
layer normalizations and DroQ networks \citep{hiraoka_dropq_2022}. Yet, the authors did not study
the effects of these choices.

\textbf{Multivariate distributional reinforcement learning} extends
distributional RL, a framework for modeling distribution over returns \citep{bellemare_distributional_2017}, 
to multi-valued reward functions. Previously, \citet{zhang_distributional_2021} modeled
the complex multivariate distribution by minimizing the maximum mean discrepancy over the 
joint returns, while \citet{wiltzer_foundations_2024} proposed a provably convergent algorithm
for learning multivariate return distributions in tabular MDPs. Instead, we propose to learn
the scalarized distributional returns directly, offering a pragmatic approach specifically 
targeted for MORL under the common linear scalarization assumption.
\citet{cai_distributional_2023} also consider the MORL setting without the linearity assumption.
They propose to model the distribution of user preferences by learning a set of plausible
scalarization functions, yet they train separate policies for each function, resulting in
poor sample efficiency.

\section{Background}\label{sec:background}

\textbf{Multi-Objective Reinforcement Learning} tasks are formally defined via Multi-objective Markov Decision Process (MOMDP) \citep{krishnendu_momdp_2006}, which is a 
tuple \(\langle \mathcal{S}, \mathcal{A}, \mathcal{P}, \mathcal{R}, \gamma, \Omega, f_\Omega \rangle\),
consisting of state-space \(\mathcal{S}\), action-space \(\mathcal{A}\),  
transition function \(\mathcal{P}: \mathcal{S} \times \mathcal{A} \times \mathcal{S} \rightarrow [0, 1]\),  
vector-valued reward function \(\mathcal{R}: \mathcal{S} \times \mathcal{A} \times \mathcal{S} \rightarrow \R^d\), 
where \(d \geq 2\) is the number of objectives, discount factor \(\gamma \in [0, 1)\), preference space \(\Omega\) 
and a scalarization function \(f_\omega: \R^d \times \Omega \rightarrow \R\).
In this paper, we limit our discussion to preference space \(\Omega = \{\pref \in \R^d \vert \sum_{i=1}^d \pref_i = 1\}\),
and to linear utility functions: \(f_\omega(\vr, \pref) = \pref^T\vr\), as is commonly done in existing MORL research
\citep{abels_dynamic_2019, kyriakis_pareto_2022, lu_capql_2023, shu_hypermorl_2024}.  As there is generally no 
policy that can maximize all objectives simultaneously, one seeks to find a set of \emph{Pareto-optimal} 
policies \citep{roijers_survey_2013}. We say that a policy \(\pi\) dominates another policy \(\pi'\) (denoted by 
\(\pi' \prec \pi\)) if \(\forall i:\;\mV^{\pi'}_i \leq \mV^{\pi}_i\) with at least one strict inequality, where
\(\mV^\pi = \E_\pi \left[\sum_{t=0}^{\infty} \gamma^t \vr_{t} \right] \in \R^d\) is vector-valued value function.
A policy is considered Pareto-optimal if there is no policy that dominates it. The set of all such policies
is called the Pareto-optimal set, and their induced value functions form the Pareto-front \citep{roijers_survey_2013}.
In this paper, we focus on \emph{general policy} methods that try to solve the above problem by finding a 
preference-conditioned policy \(\pi(\cdot \vert \vs, \pref)\) that maximizes the scalarized expected returns (SER) 
for any preference \(\pref \in \Omega\):
\begin{equation}\label{eq:ser}
    \pi^{*}_{\pref} = \argmax_{\pi \in \Pi}\; \pref^T\E_{\pi(\cdot \vert \cdot, \pref)}\left[\sum_{t=0}^{\infty} \gamma^t \vr_t \right]
\end{equation}
The final solution set is generated by conditioning the policy on a range of preferences.

\textbf{Stabilizing deep RL}:
The non-stationarity of targets in RL can lead to overfitting and unstable learning.
The existing literature has addressed this issue using several techniques. Feature normalization and 
    weight normalization prevent plasticity loss and reduce overfitting to early behaviors \citep{lee_simba_2024, lee_hyperspherical_2025}.
To prevent overfitting on high-variance features, observation normalization and feature normalization are used 
    to bring the magnitudes of hidden values in the network to a comparable scale \citep{huang_normalization_2023, lee_simba_2024}. 
Finally, distributional critic bounds gradient norms, leading to more stable learning
dynamics \citep{lee_hyperspherical_2025, palenicek_xqc_2026}. In this work, we will use the architecture from SimbaV2 \citep{lee_hyperspherical_2025}.
Namely, SimbaV2 uses (i) \(l2\) (hyperspherical) hidden feature normalization and running statistics observation normalization; (ii) the weights are normalized after every gradient update; and (iii) a distributional critic.
Additionally, the architecture features scaling layers and learnable residual connections.

\textbf{Distributional Critic}:
While classical value-based algorithms approximate the expected returns of a policy by a state-action value function $Q(s, a)$,
    the C51 algorithm \citep{bellemare_distributional_2017} extends on this idea by modeling the distribution of returns instead.
The distributions are approximated as categorical distributions over a discrete set of atoms
    $\{ z_i = V_{\min} + i \Delta z : 0 \le i < N \}, \quad  \Delta z := (V_{\max}-V_{\min})/(N-1)$
    using a parametric critic $Z_\theta(s, a)$.
Given a sample transition $(s, a, r, s')$ and next state action $a'$,
    the critic is trained by minimizing the cross-entropy loss
    $\mathcal{L}(\theta) = H(\hat{Z}(r, s'),  Z_\theta(s, a))$,
    where the target returns distribution \(\hat{Z}\) is computed by projecting the distribution 
\(r + \gamma Z(s', a')\) onto the common support.
 Given the output of the critic as probabilities 
\(p_i\) over the atoms $z_i$, the expected return can be computed as 
\(Q_\theta(s, a) = \E [Z_\theta(s, a)] = \sum_{i=0}^{N-1} p_i z_i.\)
Empirical results have shown that the distributional critic can greatly improve and stabilize learning 
\citep{bellemare_distributional_2017} and a study of the loss landscape showed that the distributional critic improves the conditioning of the problem, stabilizing the optimization \citep{palenicek_xqc_2026}.
\section{Methodology}\label{sec:methodology}
In this section, we introduce Momba, a new algorithm that integrates recent advancements in neural network design into the multi-objective setting.
We build the algorithm on top of  CAPQL \citep{lu_capql_2023}, which is an entropy-regularized algorithm akin to Soft Actor-Critic (SAC) \citep{haarnoja_soft_2018}.
This offers two benefits: First, as shown by \citet{lu_capql_2023}, the entropy bonus ensures that the induced solution sets are strictly
    convex, promoting numerically stable optimization under linear scalarization.
Second, SAC has been used as the backbone in many recent works on neural architectures in deep RL  \citep{lee_simba_2024, lee_hyperspherical_2025, palenicek_xqc_2026},
    motivating the choice of using a similar algorithm in the multi-objective case.
For the architecture, we adopt SimbaV2 \citep{lee_hyperspherical_2025}, a recent architecture for SORL, that adopts 
the three main components, (i) feature normalization, (ii) weight normalization, and (iii) distributional critic as mentioned in \Secref{sec:background}. 
\camera{
We provide a detailed description of the architectural components and normalizations in \Appref{appendix:architectural_improvements}.
}

\textbf{CAPQL:}
Consider a stochastic policy \(\pi(a \,\vert\, s, \pref)\) and let $\mathbf{G}^\pi(s_0, a_0) = \sum_{t=0}^T \gamma^t \vr(s_t, a_t) + \sum_{t=1}^T \gamma^t \mathbf{1}_dH(\pi(\cdot \mid s_t, \omega))$
    denote the entropy-augmented discounted returns of the policy $\pi$ when
    starting in state $s_0$ and selecting action $a_0$.
CAPQL trains a parametric $Q$-network $Q_\theta(s, a, \pref)$ and a policy $\pi_\phi(a \mid s, \pref)$.
The $Q$-network is trained by minimizing the mean squared error (MSE) loss
\begin{equation}
    \mathcal{L}_Q(\theta) = \E_{(s, a, \vr, s', \pref) \sim \mathcal{D}}
        \lVert \hat{\mathbf{Q}} - \mathbf{Q}_\theta(s, a, \pref) \rVert_2^2
\end{equation} 
where \(\hat{\mathbf{Q}} = \vr + \gamma ( \mQ_{\overline{\theta}}(s', a', \pref) - \alpha \mathbf{1}\log\pi_\phi(a' \mid s', \pref) ) \), \(a' \sim \pi_\phi(\cdot | s', \pref) \) denotes the target Q-value estimate, and $\overline{\theta}$ denotes the target network parameters.
Intuitively, the $Q$-value approximates 
the expected vector returns \(\E \left[\mathbf{G}^{\pi(\cdot \mid \cdot, \pref)}(s, a)\right]\).
The policy is trained for each state $s$ and preference $\pref$
to maximize the scalarized value
\begin{equation}
    \E_{a \sim \pi_\phi}[\pref^T \mathbf{Q}_\theta(s, a, \pref)] + \alpha H(\pi_\phi(\cdot \mid s, \pref)).
\end{equation}

\textbf{Multi-objective distributional critic}:
C51-style \citep{bellemare_distributional_2017} distributional critic has been a crucial component in improving the performance
    of existing deep RL algorithms \citep{lee_hyperspherical_2025, palenicek_xqc_2026}.
However, the categorical critic cannot be directly applied to model the multivariate returns distribution in MORL.
Instead, we notice that the critic predictions only affect the policy optimization via the 
scalarized Q-value \(\pref^T \mQ(s, a, \pref)\). Thus, we can avoid learning the multivariate return distribution,
and propose to use a conditioned distributional critic \(Z_\theta (s, a, \pref)\), which learns the 
univariate distribution of the scalarized returns \(\pref^T \mathbf{G}^{\pi(\cdot \vert \cdot, \pref)}(s, a)\).
This is done by minimizing the cross-entropy (CE) loss
\begin{equation}
    \mathcal{L}_Z(\theta) = \E_{(s, a, \vr, s', \pref) \sim \mathcal{D}}
        H(\hat{Z}(\pref^T \vr, s'),  Z_\theta(s, a, \pref))
\end{equation}
where \(\hat{Z}\) is the TD(0) bootstrap estimate of the scalarized returns distribution computed using the scalarized reward $\pref^T \vr$.
The actor then maximizes the value
\begin{equation}
    \E_{a \sim \pi_\phi}\left[\E[{Z}_\theta(s, a, \pref)]\right] + \alpha H(\pi_\phi(\cdot \mid s, \pref)).
\end{equation}
In order to bound  the Q-values to the support of the distributional critic,
    SORL algorithms commonly normalize the rewards by the running standard deviation of the returns
\citep{lee_hyperspherical_2025, palenicek_xqc_2026}. 
However, in MORL, the magnitude of returns can change significantly based on the
preference, and thus, we propose to normalize the rewards component-wise by the maximum
returns encountered throughout the training.  Finally, in \Appref{appendix:vector_scalar_critic} 
we describe how the distributional critic can be modeled while retaining the non-scalarized expected
returns, which are required, for instance, by envelope style algorithms \citep{yang_generalized_2019, Li_cola_2025}.

\textbf{Preference selection}: One crucial question in MORL algorithms is how the preference \(\pref\),
used as conditioning variable, is selected during training 
\citep{xu_prediction-guided_2020, hayes_practical_2022, alegre_sample-efficient_2023}. Indeed, identifying 
promising preferences can improve the convergence speed of the algorithms \citep{hayes_practical_2022, alegre_sample-efficient_2023}. As our goal is to study the effect of the architectural changes, we choose a fairly unsophisticated 
approach: We sample a new preference at the beginning of each episode from a static uniform distribution over the 
preference space. While this differs from CAPQL \citep{lu_capql_2023}, where the authors sample a new preference at each timestep,
the proposed approach is more in line with other work in MORL \citep{abels_dynamic_2019, alegre_sample-efficient_2023, xu_prediction-guided_2020, Li_cola_2025}, thus supporting our goal of studying the effect of architectural changes. We provide 
the complete pseudocode, with our changes highlighted in \Appref{appendix:pseudocode}.
\section{Experiments}\label{sec:experiments}
This section is structured as follows: Firstly, \Secref{ssec:experiment_setup} describes the experimental setup,
including the baselines and benchmarks. Then \Secref{ssec:main_experiments} evaluates whether the proposed
method results in improvements in asymptotic performance and sample efficiency. Lastly, \Secref{ssec:ablations}
performs ablations over the three key components: (i) observation and feature normalization,  (ii) weight normalization, 
and (iii) distributional critic, validating our proposed strategies.

\subsection{Experiment Setup}\label{ssec:experiment_setup}
We compare the proposed approach to 4 baselines. CAPQL \citep{lu_capql_2023} is the underlying
algorithm behind Momba, without any architectural improvements. PGMORL \citep{xu_prediction-guided_2020} 
and DPMORL \citep{cai_distributional_2023} are multi-policy methods that train separate policies for 
each scalarization function, thus leading to a discrete approximation of the Pareto-front. 
GPI-LS \citep{alegre_sample-efficient_2023} obtains best-in-class sample efficiency by 
using General Policy Improvement (GPI) for selecting preferences during training. However, this comes 
with the downside of high computational cost (300k timesteps taking more than 120 hours in our experiments).
We provide a more detailed discussion of the baselines and details on how they were run in \Appref{appendix:baselines}.

We use 7 continuous control tasks from \citet{xu_prediction-guided_2020} for evaluating our methods. 
These tasks consist of 6 environments, implemented in the MuJoCo physics 
engine \citep{todorow_mujoco_2012}: Ant, Swimmer, HalfCheetah, Humanoid, Walker2D, and Hopper.
For Hopper, we consider a two- and three-objective variant (referred to as Hopper-v4 and 
Hopper-v3, respectively). \camera{Since SimbaV2 targets continuous control domains, we omit evaluation on discrete environments. Additionally, existing discrete benchmarks for MORL consist of small state or action spaces \citep{vamplew_morl_2011, michailidis_lorenz_2026}; thus, function approximation is unlikely to be the limiting factor.
} A detailed description of the environments, including the reference points used to compute the results, is provided in \Appref{appendix:environments}.

For measuring the quality of the solution sets, we use two common metrics in MORL research, 
Hypervolume (HV) and Expected Utility Metric (EUM): 
\begin{itemize}
    \item \textbf{Hypervolume} \(\mathrm{HV}\) (\(\Uparrow\)) \citep{zitzler_hypervolume_1998} measures the space
        or volume enclosed by the solutions in the set P: 
        \(
        \mathrm{HV}(P) = \int_{\mathbb{R}^n} \mathbbm{1}_{H(P)} (z) dz
        \)
        where \(H(P) = \{z \in Z \vert \exists i: 0 \leq i \leq \vert P \vert, r_0 \preceq z \preceq P(i) \}\).
        Here \(P(i)\) is the \(i^{\mathrm{th}}\) solution,  \(r_0\) is the reference point and \(\mathbbm{1}_{H(P)}\) 
        is an indicator function. 
    \item \textbf{Expected Utility Metric} \(\mathrm{EUM}\) (\(\Uparrow\)) \citep{zintgraf_quality_2015} 
        captures the expected utility for a user from a given solution set, defined as 
        \(\mathrm{EUM}(P) = \E_{f \sim P_f} \left[\max_{\pi \in P} f(\mV^\pi)\right]\), where \(P\) is 
        the solution set and \(P_f\) is distribution over scalarization functions, which 
        simplifies to distribution over the preferences, as we are limited to linear scalarization functions. 
\end{itemize}
As noted by \citet{zintgraf_quality_2015}, hypervolume can detect improvements in uniformity, spread, and convergence of the solution set, making it a good indicator of the overall quality, and thus, we use HV as our main metric.
When reporting aggregate metrics, we normalize the HV and EUM values by the mean values Momba obtained after being trained for 1M steps. 
Regardless, we report both HV and EUM when comparing against the baselines,
while for ablations, we only report HV, as we notice that both metrics behave similarly under
linear scalarization. \camera{%
Notably, both metrics fail to explicitly capture the diversity of the generated Pareto front. Ideally, a high-quality solution set exhibits 
both high density and uniform dispersion across the objective space, thereby facilitating fine-grained 
trade-off exploration \citep{hayes_practical_2022}.  However, standard metrics, such as sparsity, rely on Euclidean
distances between solutions, rendering them ill-suited for comparing fronts with varying
convergence rates; Indeed, optimal sparsity can be trivially achieved by clustering solutions \citep{zhu_scaling_2022, liu_efficient_2025}. 
Consequently, we resort to visualizing the obtained solution sets for qualitative assessment.
}

To measure performance, we first approximate the Pareto front by averaging returns across 5 episodes for different preferences.
Following \citet{alegre_sample-efficient_2023}, we use 100 equally spaced preferences for general policy methods, while PGMORL and DPMORL are evaluated based on their current (policy, utility function) pairs. 
Finally, we compute HV and EUM from the approximated Pareto front.
When reporting results, we use interquartile mean (IQM) and stratified bootstrapped confidence intervals (SBCIs)
as recommended by \citet{agarwal_rllible_2021}.

\subsection{Experimental Validation}\label{ssec:main_experiments}
We begin by investigating whether the proposed method can improve in i) asymptotic performance and ii) sample efficiency over the baselines. \Figref{fig:main_fig}, displayed on the first page, 
answers the former question, showcasing the performance of the final solution set, aggregated over all environments. The proposed approach \emph{outperforms baselines},
achieving \camera{\(\sim 35\%\)} improvement in HV and \camera{\(\sim 16\%\)} 
improvement in EUM over the runner-up, PGMORL. When compared to CAPQL, the underlying algorithm behind Momba, 
we achieve \camera{\(\sim 132\%\)} and \camera{\(\sim 35 \%\)} improvements in aggregate HV and EUM, respectively.
However, we wish to acknowledge that the performance we report for 
DPMORL \citep{cai_distributional_2023} is worse than expected based on the original paper. 
We discuss the possible reasons for the discrepancy in DPMORL performance in \Appref{appendix:baselines}.

\begin{figure}[h]
    \centering
    \includegraphics[alt={Hypervolume (top) and EUM (bottom) as function of timesteps in Ant, Humanoid and  Hopper-v3. Plot compares the sample-efficiency of Momba, PGMORL, CAPQL and DPMORL. Momba outperforms CAPQL in all environments, while beating PGMORL in Ant and Humanoid, and matching it in Hopper-v3.},width=\linewidth]{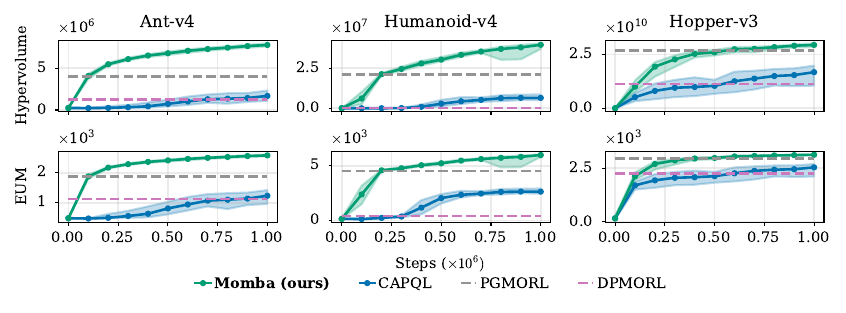}
    \caption{
        \textbf{Momba substantially improves sample efficiency against CAPQL}. The figures
        show HV (top) and EUM (bottom) (IQM and 95\% SBCIs over 10 seeds) during training. 
        Horizontal lines indicate final performance of PGMORL at \(4.8 \times 10^7\) timesteps for Ant and \(12 \times 10^7\) timesteps for Humanoid and Hopper, 
        and final performance of DPMORL at \(1 \times 10^7\) timesteps. 
    }
    \label{fig:training_curves}
\end{figure}

To examine if the proposed approach improves the sample efficiency of the underlying algorithm,
\Figref{fig:training_curves} displays the training curves of CAPQL and Momba in 3 environments.
In addition to improving on CAPQL, in Ant and Humanoid, Momba reaches \camera{the final performance of 
PGMORL} after training only for 100k and 200k steps respectively. 
We highlight that while CAPQL could not match the performance 
of PGMORL in any of the shown environments, Momba outperforms PGMORL in 2 of the 3 environments, 
and matches it in the challenging 3-objective Hopper environment, while requiring a fraction
of the training steps.

\begin{figure}[h]
    \centering
    \includegraphics[alt={Normalized HV (left) and normalized EUM (right) as a function of timesteps, aggregated over all environments with UTD 1 and 8 for Momba and UTD 1 and 20 for GPI-LS. When using similar UTD ratio, Momba beats GPI-LS in sample efficiency.},width=\linewidth]{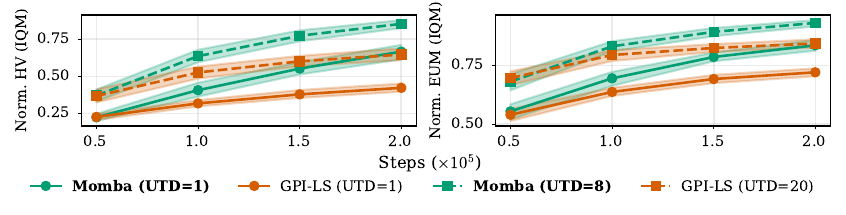}
    \caption{
        \textbf{Momba \camera{outperforms} GPI-LS at similar UTD after 200k steps}. We compare the 
        normalized HV and EUM (IQM and 95\% SBCIs over 7 seeds) of 
        Momba and GPI-LS with different UTD-ratios during the first 200k steps. 
    }
    \label{fig:gpi_comparison}
\end{figure}
To further study the sample efficiency of our method, we compare it against GPI-LS, a method 
with best-in-class sample efficiency. \Figref{fig:gpi_comparison} displays the training curves for
normalized HV and EUM with 95\% SBCIs aggregated over all environments. To remove the effect of 
update-to-data (UTD) ratio, we show the results at \(\mathrm{UTD}=\{\bm{1}, 8\}\) for Momba 
and at \(\mathrm{UTD}=\{1, \bm{20}\}\) for GPI-LS, where the default value is bolded. 
The results demonstrate that when paired with a similar UTD ratio, Momba \camera{can beat the sample efficiency of GPI-LS.}
We want to emphasize that this result is 
achieved by training Momba \emph{using preferences sampled from a static distribution}, 
whereas GPI-LS uses a sophisticated strategy for preference selection, 
highlighting the importance of an expressive neural network.

\begin{figure}[htbp]
    \centering
    \includegraphics[alt={Generated solution sets in Walker2d (left), HalfCheetah (middle) and Hopper (right). In Walker2d and Hopper, Momba generates superior solution set, while in HalfCheetah the solution set has narrower coverage. There also exists some gaps in upper left corner of both Walker2d and HalfCheetah, likely due to linear scalarization.},width=\textwidth]{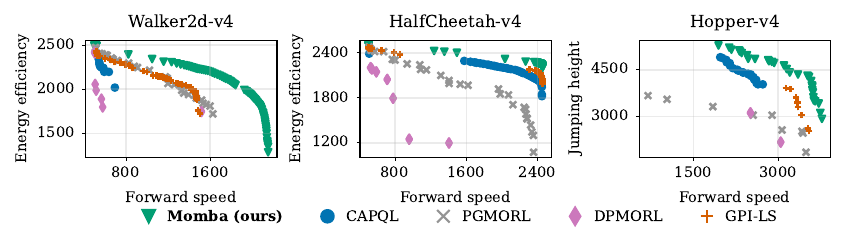}
    \caption{
        \textbf{Momba produces solution sets with good coverage}.
        We visualize the generated solution sets in Walker2d (left), HalfCheetah (middle), and Hopper (right).
        Momba produces superb solution sets in Walker2d and Hopper. 
        Gaps in the upper left corner of Walker2d and HalfCheetah are 
        likely a result of poor numerical properties caused by linear scalarization.
    }
    \label{fig:pf_visualization}
\end{figure}

Lastly, we display the generated solution sets in 3 environments in \Figref{fig:pf_visualization} for qualitative 
analysis. In Walker2d and Hopper-v4, Momba produces solution sets with good coverage, outperforming other 
methods. In HalfCheetah, Momba has a more limited coverage, while still producing a solution set that dominates
most of the baselines. We note that in Walker2d and HalfCheetah, the upper left corner contains gaps, 
which we attribute to our use of linear scalarization. While entropy regularization turns the set of 
induced value functions strictly convex \citep{lu_capql_2023}, it is possible that a stronger
regularization would be required to reliably obtain policies in these regions.

\subsection{Validating design choices}\label{ssec:ablations}
\begin{wrapfigure}{r}{0.4\textwidth}
    \centering
    \includegraphics[alt={Normalized hypervolume as a function of critic parameters with MLP or Simba architecture and MSE or CE critic loss. While Simba+CE is the most effective approach, pairing  MLP with  the CE loss is surprisingly effective, outperforming Simba with MSE loss when the critic size is increased.},width=\linewidth]{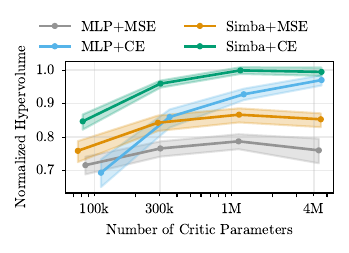}
    \caption{
        \textbf{The categorical critic adaptation is highly effective}. The figure displays
        normalized HV (IQM and 95\% SBCIs over 10 seeds) as a function of the critic size.
        Momba is Simba+CE. 
    }
    \label{fig:critic_scaling}
\end{wrapfigure}

In this section, we explore the effectiveness of our three key components: (i) distributional critic, 
(ii) feature and observation normalization, and (iii) weight normalization. We begin by comparing the distributional 
critic with the cross-entropy loss (CE) to the non-distributional critic with standard mean squared error loss (MSE).
To ensure that any performance improvements are not due to 
architectural changes introduced in Simba, such as normalizations and residual connections, we test the performance 
of the proposed distributional critic with the standard 2-layer MLP, commonly used to represent the Q-network
in MORL algorithms. In order to eliminate the effect of the size of the critic, we vary the hidden dimension from 
256 to 2048 and from 64 to 512 for MLP and Simba, respectively, resulting in networks with approximately
similar parameter sizes. 
\Figref{fig:critic_scaling} showcases that the proposed approach (Momba=Simba+CE) 
clearly outperforms the Simba+MSE variant across all critic sizes. Perhaps surprisingly, the standard
MLP with categorical loss (MLP+CE) \emph{matches or even outperforms Simba+MSE}, showcasing the
effectiveness of the proposed categorical critic adaptation to the multi-objective domain. 
\camera{
    The default MLP with MSE loss corresponds to a parameter-scaled version of CAPQL, the base algorithm
    behind Momba.
}
\camera{As expected, this combination retains poor performance, regardless of the critic
    size, implying that the performance gains are \emph{not due to the increase in the parameter counts}, but rather stem from the 
    improvements to function approximation and critic loss.
}

\begin{figure}[htbp]
    \centering
    \includegraphics[alt={Left: Normalized hypervolume with all possible combinations of feature, weight and observation normalization. Combination of all normalization yields the best performance. Right: Shapley values of the normalizations. Observation normalization is given the highest importance, while feature and weight normalization have similar importance.},width=\linewidth]{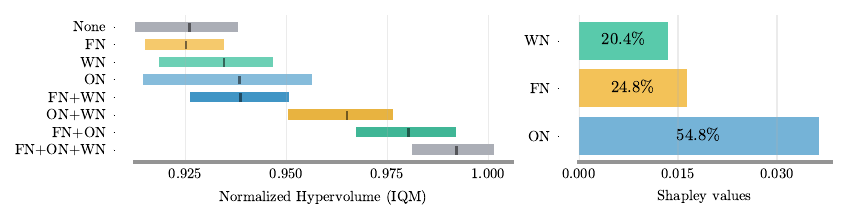}
    \caption{
        \textbf{Observation normalization is the most effective component}. Left: We investigate
        the effect of i) Feature Normalization (FN), ii) Observation normalization (ON), and iii)
        Weight Normalization (WN) to the final HV (IQM and 95\% SBCIs over 10 seeds). Momba is FN+ON+WN.
        Right: Observation normalization is associated with the highest Shapley value, while 
        weight and feature normalization are given similar importance. 
    }
    \label{fig:norm_ablations}
\end{figure}

To validate the effect of the (i) weight normalization (WN), (ii) feature normalization (FN), 
and (iii) observation normalization (ON), we evaluate Momba under all combinations of these 
normalization methods. \Figref{fig:norm_ablations} displays the normalized HV and 95\% SBCIs over 10 seeds, 
aggregated over all tasks, and the Shapley values of the three different normalization techniques. 

The results demonstrate that observation normalization was the most effective technique on our benchmarks,
accounting for \camera{\(\approx 55\%\)} of the improvements according to the Shapley values. 
While feature and weight normalization provide more modest improvements, 
all normalization schemes synergistically improve the performance.

\section{Conclusions \& Discussions}\label{sec:discussions}
In this paper, we investigated whether the recent advances in neural network design, such as i) feature \& observation 
normalization, ii) weight normalization, and iii) distributional critic, 
\camera{
can improve the performance of an existing MORL algorithm. 
We applied and adapted a recent deep RL architecture, SimbaV2, to an existing entropy-regularized MORL algorithm,
which we kept close to the original, only changing the preference sampling rate and critic loss.
} To deal with the multivariate return distribution, we proposed a simple yet effective tactic of modeling scalarized 
return distribution, motivated by the prevalent scalarized expected return objective. Our extensive experiments
demonstrated that one can achieve competitive performance while matching or improving the sample efficiency of the base algorithm \camera{in multi-objective continuous control tasks}. We demonstrated the effectiveness of our adaptations via ablations.

We acknowledge that our study has several limitations. First, we restrict attention to linear scalarization. 
While common in MORL, nonlinear or learned scalarization may be preferable in some 
settings \citep{hayes_practical_2022,cai_distributional_2023}. Notably, our distributional critic 
does not assume a specific scalarization form; the constraint arises from TD learning via the Bellman error, 
which may not hold under nonlinear scalarization \citep{roijers_multi-objective_2018,hayes_practical_2022}. 
\camera{
Second, our benchmarks are limited to continuous control tasks, with 2 objectives, with only one with 3 
objectives. This makes it difficult to assess how applicable the proposed changes would be in
discrete state and action spaces, and how well the approach would scale with increasing number of 
objectives.}
Finally, our more expressive function approximators can increase wall-clock time, though recent systems 
advances (e.g., JIT compilation) can mitigate this \citep{frostig_jax_2019,ansel_torch_2024},  and replacing 
SimbaV2 with the XQC architecture may offer similar gains with simpler designs \citep{palenicek_xqc_2026}.



\appendix

%
%
%

\subsubsection*{Acknowledgments}
\label{sec:ack}
\camera{
Adam Štafa and Petr Novotný were supported by U.S. Army Research Office under 
contract number W911NF261A180.
Santeri Heiskanen was supported by the Ministry of Education and Culture’s Doctoral Education Pilot in Finland 
(Decision No. VN/3137/2024-OKM-6; Finnish Doctoral Program Network in Artificial Intelligence, AI-DOC). 
We acknowledge the computational resources provided by the Aalto Science-IT project.
}

%

\bibliography{main}
\bibliographystyle{rlj}

\beginSupplementaryMaterials

\section{Full Evaluation Results}\label{appendix:full_evaluation_results}

\camera{
To facilitate more detailed analysis of the proposed method, we report the performance metrics of Momba and baselines 
across all environments in \Tabref{tab:full_results}. The results showcase that Momba can outperform all the baselines in most of the
continuous control tasks, with the exception of Swimmer, where PGMORL outperforms Momba. However, this environment is clearly difficult 
for general policy methods, likely due to high imbalance between the objective difficulties. It should be noted that in the rest of the
environments, Momba's and the runner-up methods 95\% SBCIs for Hypervolume don't overlap. While GPI-LS is trained for considerably fewer timesteps,
this matches the default training protocol used in the original paper. Moreover, as is mentioned in \Secref{ssec:experiment_setup}, training
GPI-LS for more timesteps is extremely time-consuming. The hyperparameters for Momba and a detailed description of the baselines are given in \Appref{appendix:training_details} and 
\Appref{appendix:baselines} respectively.
}

\begin{table}[ht!]
\small
\centering
\caption{
\camera{
Full results of our evaluation. We report the IQM and \(95\%\) SBCIs of hypervolume (\(\Uparrow\)) and EUM (\(\Uparrow\)) across 10 seeds 
for each algorithm and environment. The best and runner-up methods for each environment are \textbf{bolded} and \underline{underlined} respectively.
}
}
\label{tab:full_results}
\tablesize
\camera{
\begin{tabular}{@{}llccccc@{}}
\toprule
\textbf{Environment} & \textbf{Metric} & \textbf{\textsc{Momba}} & \textbf{\textsc{PGMORL}} & \textbf{\textsc{GPI-LS}} & \textbf{\textsc{CAPQL}} & \textbf{\textsc{DPMORL}} \\
\midrule
\multirow{3}{*}{\textsc{Ant-v4}} 
& Steps (\(\times 10^6\)) & 1 & 48 & 0.20 & 1 & 10 \\ 
\cmidrule(lr){2-7}
& HV ($\times 10^{6}$) & \textbf{7.78} {\cisize \textcolor{black!60}{[7.61, 7.97]}} & 4.01 {\cisize \textcolor{black!60}{[3.16, 4.57]}} & \underline{4.23} {\cisize \textcolor{black!60}{[3.73, 4.49]}} & 1.69 {\cisize \textcolor{black!60}{[1.08, 2.33]}} & 1.28 {\cisize \textcolor{black!60}{[1.17, 1.40]}} \\
\cmidrule(lr){2-7}
 & EUM ($\times 10^{3}$) & \textbf{2.58} {\cisize \textcolor{black!60}{[2.55, 2.61]}} & 1.88 {\cisize \textcolor{black!60}{[1.66, 2.00]}} & \underline{1.93} {\cisize \textcolor{black!60}{[1.81, 1.99]}} & 1.23 {\cisize \textcolor{black!60}{[0.96, 1.42]}} & 1.13 {\cisize \textcolor{black!60}{[1.07, 1.19]}} \\
\midrule
\multirow{3}{*}{\textsc{HalfCheetah-v4}} 
& Steps (\(\times 10^6\)) & 1 & 30 & 0.20 & 1 & 10 \\ 
\cmidrule(lr){2-7}
& HV ($\times 10^{6}$) & \textbf{5.84} {\cisize \textcolor{black!60}{[5.79, 5.87]}} & 4.86 {\cisize \textcolor{black!60}{[4.58, 5.01]}} & \underline{5.53} {\cisize \textcolor{black!60}{[4.70, 5.61]}} & 5.42 {\cisize \textcolor{black!60}{[4.20, 5.52]}} & 2.39 {\cisize \textcolor{black!60}{[2.31, 2.46]}} \\
\cmidrule(lr){2-7}
 & EUM ($\times 10^{3}$) & \textbf{2.34} {\cisize \textcolor{black!60}{[2.33, 2.35]}} & 2.07 {\cisize \textcolor{black!60}{[2.02, 2.11]}} & \underline{2.30} {\cisize \textcolor{black!60}{[2.10, 2.33]}} & 2.23 {\cisize \textcolor{black!60}{[1.99, 2.26]}} & 1.56 {\cisize \textcolor{black!60}{[1.54, 1.58]}} \\
\midrule
\multirow{3}{*}{\textsc{Hopper-v3}} 
& Steps (\(\times 10^6\)) & 1 & 120 & 0.20 & 1 & 10 \\ 
\cmidrule(lr){2-7}
& HV ($\times 10^{10}$) & \textbf{2.95} {\cisize \textcolor{black!60}{[2.85, 3.02]}} & \underline{2.68} {\cisize \textcolor{black!60}{[2.62, 2.80]}} & 2.07 {\cisize \textcolor{black!60}{[1.72, 2.23]}} & 1.68 {\cisize \textcolor{black!60}{[1.05, 1.99]}} & 1.12 {\cisize \textcolor{black!60}{[0.96, 1.30]}} \\
\cmidrule(lr){2-7}
 & EUM ($\times 10^{3}$) & \textbf{3.12} {\cisize \textcolor{black!60}{[3.09, 3.13]}} & \underline{2.95} {\cisize \textcolor{black!60}{[2.92, 3.00]}} & 2.81 {\cisize \textcolor{black!60}{[2.65, 2.84]}} & 2.53 {\cisize \textcolor{black!60}{[2.10, 2.69]}} & 2.25 {\cisize \textcolor{black!60}{[2.12, 2.38]}} \\
\midrule
\multirow{3}{*}{\textsc{Hopper-v4}} 
& Steps (\(\times 10^6\)) & 1 & 48 & 0.20 & 1 & 10 \\ 
\cmidrule(lr){2-7}
& HV ($\times 10^{7}$) & \textbf{1.81} {\cisize \textcolor{black!60}{[1.78, 1.85]}} & \underline{1.53} {\cisize \textcolor{black!60}{[1.40, 1.64]}} & 1.35 {\cisize \textcolor{black!60}{[1.29, 1.42]}} & 1.21 {\cisize \textcolor{black!60}{[1.11, 1.30]}} & 0.88 {\cisize \textcolor{black!60}{[0.75, 1.00]}} \\
\cmidrule(lr){2-7}
 & EUM ($\times 10^{3}$) & \textbf{4.07} {\cisize \textcolor{black!60}{[4.03, 4.10]}} & \underline{3.76} {\cisize \textcolor{black!60}{[3.60, 3.90]}} & 3.55 {\cisize \textcolor{black!60}{[3.48, 3.63]}} & 3.43 {\cisize \textcolor{black!60}{[3.27, 3.54]}} & 2.92 {\cisize \textcolor{black!60}{[2.70, 3.10]}} \\
\midrule
\multirow{3}{*}{\textsc{Humanoid-v4}} 
& Steps (\(\times 10^6\)) & 1 & 120 & 0.20 & 1 & 10 \\ 
\cmidrule(lr){2-7}
& HV ($\times 10^{7}$) & \textbf{3.96} {\cisize \textcolor{black!60}{[3.69, 4.05]}} & \underline{2.12} {\cisize \textcolor{black!60}{[2.02, 2.27]}} & 0.00 {\cisize \textcolor{black!60}{[0.00, 0.00]}} & 0.65 {\cisize \textcolor{black!60}{[0.51, 0.86]}} & 0.02 {\cisize \textcolor{black!60}{[0.01, 0.02]}} \\
\cmidrule(lr){2-7}
 & EUM ($\times 10^{3}$) & \textbf{5.99} {\cisize \textcolor{black!60}{[5.81, 6.05]}} & \underline{4.54} {\cisize \textcolor{black!60}{[4.48, 4.63]}} & -0.01 {\cisize \textcolor{black!60}{[-0.02, 0.16]}} & 2.65 {\cisize \textcolor{black!60}{[2.47, 2.94]}} & 0.41 {\cisize \textcolor{black!60}{[0.39, 0.46]}} \\
\midrule
\multirow{3}{*}{\textsc{Swimmer-v4}} 
& Steps (\(\times 10^6\)) & 1 & 12 & 0.20 & 1 & 10 \\ 
\cmidrule(lr){2-7}
& HV ($\times 10^{4}$) & \underline{1.66} {\cisize \textcolor{black!60}{[1.57, 1.74]}} & \textbf{1.75} {\cisize \textcolor{black!60}{[1.10, 2.29]}} & 0.75 {\cisize \textcolor{black!60}{[0.73, 0.98]}} & 0.56 {\cisize \textcolor{black!60}{[0.52, 0.61]}} & 1.20 {\cisize \textcolor{black!60}{[1.17, 1.31]}} \\
\cmidrule(lr){2-7}
 & EUM ($\times 10^{2}$) & \underline{1.23} {\cisize \textcolor{black!60}{[1.21, 1.25]}} & \textbf{1.27} {\cisize \textcolor{black!60}{[1.07, 1.45]}} & 0.99 {\cisize \textcolor{black!60}{[0.98, 1.04]}} & 0.93 {\cisize \textcolor{black!60}{[0.92, 0.94]}} & 1.10 {\cisize \textcolor{black!60}{[1.09, 1.13]}} \\
\midrule
\multirow{3}{*}{\textsc{Walker2d-v4}} 
& Steps (\(\times 10^6\)) & 1 & 30 & 0.20 & 1 & 10 \\ 
\cmidrule(lr){2-7}
& HV ($\times 10^{6}$) & \textbf{5.15} {\cisize \textcolor{black!60}{[4.17, 5.32]}} & \underline{3.47} {\cisize \textcolor{black!60}{[3.23, 3.64]}} & 3.33 {\cisize \textcolor{black!60}{[3.03, 3.67]}} & 2.38 {\cisize \textcolor{black!60}{[1.82, 2.66]}} & 2.54 {\cisize \textcolor{black!60}{[2.33, 2.70]}} \\
\cmidrule(lr){2-7}
 & EUM ($\times 10^{3}$) & \textbf{2.12} {\cisize \textcolor{black!60}{[1.76, 2.15]}} & \underline{1.80} {\cisize \textcolor{black!60}{[1.77, 1.83]}} & 1.78 {\cisize \textcolor{black!60}{[1.72, 1.84]}} & 1.62 {\cisize \textcolor{black!60}{[1.54, 1.67]}} & 1.66 {\cisize \textcolor{black!60}{[1.61, 1.70]}} \\
\bottomrule
\end{tabular}
}
\end{table}

\section{Vectorized Versus Scalarized Returns}\label{appendix:vector_scalar_critic}
In \Secref{sec:methodology}, we propose to directly learn the scalarized distributional returns.
However, in MORL literature, it is customary to predict the vector returns, which allows algorithms 
to take advantage of this information, for e.g., envelope-style updates \citep{yang_generalized_2019, Li_cola_2025}.
We show that it is possible to learn the vector returns with the distributional critic by approximating 
the marginals of the returns distributions, from which one can recover the vector Q-values by taking component-wise
expectation of the critic output.

To achieve this, we implemented the vector distributional critic as \(d\) distributional critics 
\(Z_\theta^1, \dots,  Z_\theta^d\), each predicting the distribution of one component of the \(d\)-dimensional returns.
The critic loss is
\begin{equation}
    \mathcal{L}_Z(\theta) = \E_{(s, a, \vr, s', \pref) \sim \mathcal{D}}
            \sum_{i=1}^d H(\hat{Z}_i,  Z_\theta^i(s, a, \pref)).
\end{equation}
where $\hat{Z}_i$ is computed from the $i$-th component of the reward $\vr_i$.
The actor then maximizes the value
\begin{equation}
    \E_{a \sim \pi_\phi}\left[\sum_{i=1}^d \pref_i\E\left[{Z}_\theta^i(s, a, \pref)\right]\right] + \alpha H(\pi_\phi(\cdot \mid s, \pref)).
\end{equation}
In practice, the vector critic is implemented as separate output layers, thus sharing most of their parameters.
Analogously to CAPQL, we implement clipped double Q-learning by taking argmin of the scalarized (non-distributional) value.

\begin{figure}[htbp]
    \centering
    \includegraphics[alt={Normalized hypervolume as bar plots using vectorized and scalarized critics either with MLP or Simba architecture and MSE or CE loss. In all cases, the performance between the vectorized and scalarized approach is similar.},width=\linewidth]{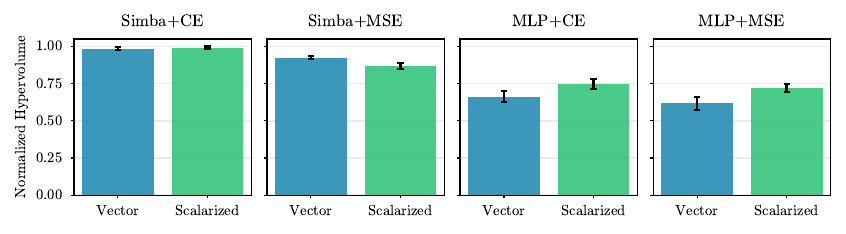}
    
    \caption{
        Normalized HV (IQM and 95\% SBCIs from 10 seeds), aggregated over all environments, 
        with all proposed components. Learning the distribution of each reward component separately
        matches the performance of learning the scalarized return distribution. Momba is Simba+CE.
    }
    \label{fig:vector_ablations}
\end{figure}

We compared the vector critic to the scalarized critic, used in the main text, 
in \Figref{fig:vector_ablations}. Similarly to \Secref{ssec:ablations}, the comparison is performed with
Simba and MLP architecture. For completeness, we also report the performance of the non-distributional vector critic.
As can be seen in \Figref{fig:vector_ablations}, the performance of the two approaches is nearly identical, 
demonstrating that our proposed adaptation of the distributional critic could also be considered in cases where the 
vector-values are required.
\section{Detailed Description of Architectural Improvements}\label{appendix:architectural_improvements}

\camera{
As mentioned in Section \ref{sec:background}, SimbaV2 architecture 
\citep{lee_hyperspherical_2025} consists of three main components, which we describe in detail here.}

\textbf{Observation and feature normalization}: We follow \citet{lee_hyperspherical_2025}, and apply
Running Statistic Normalization (RSNorm) to normalize each dimension of the observations \(\vo\)
to zero mean and unit variance. More specifically, at each timestep \(t\), the running mean 
\(\bm{\mu}_t \in \R^{\left|O\right|}\) and variance \(\bm{\sigma}^2 \in \R^{\left|O\right|}\)
are updated with the standard online algorithm
\begin{equation}
    \bm{\mu}_t  = \bm{\mu}_{t-1} + \frac{1}{t} \bm{\delta}_t, \quad
    \bm{\sigma}^2_t  = \bm{\sigma}^2_{t-1} + \frac{1}{t}(\bm{\delta}^2_t - \bm{\sigma}^2_{t-1}) 
\end{equation}
where \(\bm{\delta}_t= \vo_t - \bm{\mu}_{t-1}\). Using these running statistics, a given 
observation is then normalized as 
\begin{equation}
    \overline{\vo}_t = \mathrm{RSNorm}(\vo_t) = \frac{\vo_t - \bm{\mu}_t}{\sqrt{\bm{\sigma}^2_t + \epsilon}}
\end{equation}
The normalized observation \(\overline{\vo}\) is then mapped onto a unit hypersphere to stabilize learning. To ensure that the magnitude information is not lost during the mapping, \citet{lee_hyperspherical_2025} propose to concatenate a
positive constant \(c_{\mathrm{shift}}\) to the observation vector before mapping the vector. Then, the final normalized
observation becomes 
\begin{equation}
    \tilde{\vo}_t = \mathrm{L2norm}([\overline{\vo}_t, c_{\mathrm{shift}}])
\end{equation}
where \(\mathrm{L2norm}\) divides each component of the vector by the square root of sum
of squares of the vector, and \([\cdot, \cdot]\) denotes the concatenation operation.

\textbf{Architecture}: SimbaV2 introduces two new formulations for the learnable layers. 
Firstly, instead of using standard linear layers, \citet{lee_hyperspherical_2025} decompose 
the standard linear layer into a linear layer (without a bias) with weights that are
constrained to the unit hypersphere, and a learnable scaling vector that is applied element-wise.
To ensure that the weights remain in the hypersphere, \(\mathrm{L2Norm}\) is applied after each 
gradient update. Secondly, instead of using traditional skip-connections, the authors propose the
use of learnable interpolation layers (LERP), which can interpolate between the original input
\(\vh^l\) and non-linearly transformed input \(\tilde{\vh}^l\):
\begin{equation}
    \vh^{l+1} = \mathrm{L2Norm}((1 - \alpha) \odot \vh_t^l + \alpha \odot \tilde{\vh}^l_t)
\end{equation}
where \(\alpha\) is a learnable parameter.

\textbf{Reward scaling}: as noted in \Secref{sec:methodology}, the C51-style categorical critic 
requires one to map the returns to interval \([G_{\mathrm{min}}, G_{\mathrm{max}}]\). 
We follow SimbaV2 \citep{lee_hyperspherical_2025}, and achieve this by applying the normalization
to rewards directly. In addition to limiting the returns to a given interval, this also 
i) provides stable gradient for both actor and critic, and ii) ensures that the reward
components have similar magnitudes, which is crucial when applying reward scalarization. In
detail, we normalize each reward component based on the observed maximum returns so far:
\begin{equation}
    \tilde{R}^i_t = v_{\mathrm{max}} \cdot \frac{R_t^i}{G_{t, \mathrm{max}}^i}, \quad i=1, \dots d
\end{equation}
where \(v_{\mathrm{max}} \in \R\) is the desired normalized return, \(R^i_{t}\)
is the \(i\mathrm{th}\) reward component and \(G^i_{t, \mathrm{max}}\) is the maximum observed 
return for the \(i\mathrm{th}\) reward component. Intuitively, this normalization simply 
ensures that the normalized returns are in the symmetric range 
\([-v_{\mathrm{max}}, v_{\mathrm{max}}]\). SimbaV2 also included an additional term to 
ensure that the variance of the rewards is close to 1, but we found that to bring no benefits, and
thus excluded it.
\section{Training Details}\label{appendix:training_details}
We report the default hyperparameters used in our experiments in \Tabref{tab:training_details}. 
We did not perform extensive hyperparameter optimization, and thus, it is possible that better performance
could be extracted via large-scale hyperparameter search. Best return scale denotes the coefficient used 
to compute the scaling factor for rewards. Notably, we automatically tune the entropy
regularization coefficient, while the original CAPQL implementation used fixed entropy coefficients, tuned 
specifically for each environment.

\begin{table}[ht!]
\small
\centering
\caption{Default hyperparameters for Momba, used in our main experiments}
\label{tab:training_details}
\begin{tabular}{l l | l l}
\toprule
\textbf{Parameter} & \textbf{Value(s)} & \textbf{Parameter} & \textbf{Value(s)} \\
\midrule
Training steps  & 1M & Actor blocks & 2 \\
Replay buffer size & 1M & Actor hidden dim & 128 \\
Initial training steps & 500 & Critic blocks & 2  \\
$\gamma$ & 0.99 & Critic hidden dim & 256 \\
Target critic momentum & 0.05 & Number of critics &  2 \\
Optimizer & Adam & Number of atoms & 101 \\
Policy LR & \(1 \times 10^{-4}\) & Best return scale & 3.0 \\
Critic LR &  \(1 \times 10^{-4}\) & Categorical support & \([-5, 5]\)  \\
Initial \(\alpha\)  & 0.2  & UTD & 1 \\
Target \(\alpha\) & \(-\lvert A \lvert\) & Batch size & 256 \\

\bottomrule
\end{tabular}
\end{table}

\section{Algorithmic Modifications}\label{appendix:pseudocode}
As noted in the main text, while our goal was to avoid making algorithmic changes, we ended up doing \emph{three}
changes to the original CAPQL \citep{lu_capql_2023}  to make our results more broadly
applicable. Firstly, instead of using normal distribution as our preference sampling
distribution, as done by \citet{lu_capql_2023}, we opt to use Dirichlet distribution \(\mathrm{Dir}(\bm{1}_k)\)
instead. This choice was made mostly due to convenience, as the Dirichlet distribution has exactly the same support
as our preference space \(\Omega\). Secondly, we sample a new preference once per episode (Line 4, \Algref{alg:capql}),
as opposed to sampling a new preference for each timestep. This change was done, since this is a far more common
choice in MORL algorithms \citep{abels_dynamic_2019, xu_prediction-guided_2020, alegre_sample-efficient_2023, Li_cola_2025}. 
Lastly, we update the critic loss function (Line 11 in \Algref{alg:capql})
to use the categorical loss, similar to \citet{bellemare_distributional_2017}. As discussed in the main text,
using categorical loss for the critic is one of the main components driving the improved performance of the
proposed architecture. The complete algorithm, with the proposed changes highlighted, is displayed in
\Algref{alg:capql}.

\begin{algorithm}[htb]
      \DontPrintSemicolon
      \SetAlgoLined
      \SetAlgoVlined
      \SetKwInOut{Input}{Input}\SetKwInOut{Output}{output}
      \Input{\blue{preference sampling distribution \(D_\psi = \mathrm{Dir}(\bm{1}_d)\)}}
      Initialize parameter vectors \(\theta_1, \theta_2, \overline{\theta}_1, \overline{\theta}_2, \phi\)\;
      \(\overline{\theta}_i \leftarrow \theta_i\) for \(i \in \{1, 2\}\)\;
      \ForEach{iteration}
      {
        \blue{Sample \(\pref \sim D_\psi\)} \tcp*{Sample preference once per episode}
         \ForEach{environment step}{
            \(a_t \sim \pi_\phi(s_t, \pref)\)\;
            \(s_{t+1} \sim \mathcal{P}(a_t, s_t)\)\;
            \(\mathcal{D} \leftarrow \mathcal{D}\cup\{(s_t, a_t, \mathcal{R}(a_t, s_t), s_{t+1}, \pref)\}\)\;
         }

         \ForEach{training step}{
            \(\mathcal{S} \leftarrow\) sample \(N\) transitions from \(\mathcal{D}\)\;
            \tcc{Update Critic networks using categorical loss}
            \blue{%
            \(\theta_i \leftarrow \theta_i - \lambda_\theta \nabla_{\theta_i}(
               \mathbb{E}_{\mathcal{S}}H(\hat{Z}(s_t, a_t, \pref),
               Z_{\theta_i}(s_t, a_t, \pref)
               ))\)} for \(i \in \{1, 2\}\)\;
               \Indp Where \(
                  \hat{Z}(s_t, a_t, \pref) = \pref^T \mathcal{R}(a_t, s_t) 
                  + \gamma(\min_{i\in\{1, 2\}}Z_{\overline{\theta}_i}(s_{t+1}, a_{t+1}, \pref) 
                  - \alpha \log\; \pi_\phi(a_{t+1}, s_{t+1}, \pref)\mathbf{1})
               \) and \(a_{t+1} \sim \pi_{\phi}(s_{t+1}, \pref)\)\;
               \Indm
               \tcc{Update the policy parameters}
               \(\phi \leftarrow \phi - \lambda_\pi \nabla_\phi \E_{\mathcal{S}}\left(D_{\mathrm{KL}}(\pi_\phi(\cdot, s_j, \pref) \| \frac{\exp(\pref^T \min_{i \in \{1, 2\}} Q_{\theta_i}(s_j, \pref)/\alpha)}{\Delta(s_j, \pref)})\right)\)\;
               \Indp with \(\Delta(s_j, \pref) = \int_{\mathcal{A}} \exp(\pref^T \min_{\{1, 2\}} Q_{\theta_i}(s_j, a, \pref)/\alpha)\, da\)\;
              \Indm
         \(\overline{\theta}_i \leftarrow \tau \theta_i + (1 - \tau)\overline{\theta}_i\;\mathrm{for}\;i\in \{1, 2\}\)
         }
      }
   \caption{CAPQL algorithm \cite{lu_capql_2023} with the modifications highlighted in \blue{blue}}
   \label{alg:capql}
\end{algorithm}
\section{Baselines}\label{appendix:baselines}

In this section, we give a brief description of the baselines used in this work, and how
the results showcased in the main text were obtained.

\textbf{CAPQL} \citep{lu_capql_2023} is a general policy method that also acts as a 
backbone of our experiments. It utilizes an entropy regularized formulation, similar
to Soft-Actor Critic (SAC) \citep{haarnoja_soft_2018} for learning an approximation 
of the solution set. However, the authors' motivation differs from SAC: They showcase
that by utilizing entropy regularization, one can ensure that the solution set becomes 
strictly convex, making it possible to retrieve flat portions of the Pareto front
even with linear scalarization. In our experiments, we used an implementation from 
MORL-baselines \citep{felten_toolkit_2023}, \camera{utilizing a 2 layer MLP for representing 
the policy and value-functions} and the algorithm was trained for \(1 \times 10^6\) 
timesteps.

\textbf{GPI-LS} \citep{alegre_sample-efficient_2023} is a recent 
MORL method that, similarly to CAPQL, learns a single policy that is conditioned on 
the desired preference. The algorithm prioritizes preferences during training based on the 
magnitude of achievable improvement via General Policy Improvement (GPI). 
The novel GPI-based algorithm leads to best-in-class 
sample efficiency, yet the runtime of the method is quite high, as 
the process for evaluating the GPI policy is time-consuming. In this paper, we use the original 
implementation from MORL-baselines \citep{felten_toolkit_2023}. \camera{
Notably, this implementation uses layer normalizations and DroQ networks \citep{hiraoka_dropq_2022}, making it 
the only baseline to utilize more recent architectural improvements.
}
We train the algorithms for 200k timesteps, as this is close to the default setting
used in the original paper. We also tried training the algorithms
for 300k timesteps, but in this case, the experiments could not finish in 5 days (120 hours).
We also note that GPI-LS performs \emph{additional environment steps}
to evaluate the policy at corner weights (refer to Algorithm 1 in \citep{alegre_sample-efficient_2023})
during the GPI update. To the best of our knowledge, these additional steps \emph{\textbf{are not}} counted towards
the timesteps reported in the training curves.

\textbf{PGMORL} \citep{xu_prediction-guided_2020} represents a multi-policy
method that trains separate neural networks for different preferences. The
authors evolve a population of policies using an evolutionary algorithm, which
is paired with a prediction model that tries to detect which preferences
will improve the current solution set the most. The individual policies are 
trained using multi-objective policy gradient. We utilize the original
PGMORL implementation in our experiments, using the original hyperparameters and \camera{two-layer MLPs for representing policies}. 
With these parameters, the method uses \(1.2 \times 10^7\) to \(1.2 \times 10^8\)
timesteps, depending on the environment. 

\textbf{DPMORL} \citep{cai_distributional_2023} is a multi-policy method 
that firstly learns a set of plausible utility functions using a diversity-based objective,
and then trains separate policies for each learned utility function. The authors
also modify the objective for the policy optimization to take into account distributed
returns, thus offering a direct comparison with our simpler approach for modeling
the distributional returns. In our experiments, we utilize the original implementation
by the authors. We use the default hyperparameters from the original paper, and thus,
in each environment, we train at most 10 policies, \camera{each represented using a 3 layer MLP}. In total, 
these policies are trained for \(1 \times 10^7\) timesteps. 

We would like to note that in our evaluations (presented in Figures \ref{fig:main_fig}, 
\ref{fig:training_curves} and \ref{fig:pf_visualization}), DPMORL performs 
worse than expected based on the results in the original paper. We hypothesize possible causes
for this discrepancy. Firstly, the author-provided code seems to differ from the description in the 
paper, as it doesn't seem to augment the state space with the cumulative multivariate returns 
(refer to Algorithm 1 in \citet{cai_distributional_2023}). While we tried enabling the 
state-space augmentation, it resulted in even further performance deterioration. Secondly,
the original code utilizes environment-specific precomputed values for normalizing the utility 
functions, which we do not have access to, as our environments differ from the original setup.
To the best of our knowledge, the method for computing these normalization values is not mentioned in the
original paper (this normalization seems to differ from the one mentioned in Appendix B in  \citet{cai_distributional_2023}), making it challenging for us to reproduce the results.

\camera{
For all evaluated algorithms, we summarize the number of trainable parameters and the number of training steps in \Tabref{tab:algo_parameter_counts}.
While Momba uses more trainable parameters than the baselines, its superior performance cannot be attributed solely to the number of parameters.
In \Figref{fig:critic_scaling}, we showed that distributional value representation and the architectural components are necessary to enable parameter scaling.
}

\begin{table}[htbp]
\centering
\caption{
    \camera{
        \textbf{Summary of evaluated algorithms.} We report the number of trainable parameters for the policy and value networks and the number of training steps. In multi-policy methods, the parameters are for a single policy, and the training steps are shared between the policies.
    }}
    \label{tab:algo_parameter_counts}
    \camera{
        \begin{tabular}{cccrc}
        \toprule
        Algorithm & Classification & Policies & Trainable parameters & Training steps\\
        \midrule
        Momba & General policy & 1 & 2,500,805 & 1M\\
        CAPQL & General policy & 1 & 258,723 & 1M\\
        GPI-LS & General policy & 1 & 259,119 & 200k\\
        DPMORL & Multi-policy & 10 & 17,477 & 10M\\
        PGMORL & Multi-policy & 6--15 & 17,551 & 12M--120M\\
        \bottomrule
        \end{tabular}
    }
\end{table}

\section{Environments}\label{appendix:environments}
In this section, we provide a brief overview of the environments used in this work. 
They are originally from \citep{xu_prediction-guided_2020}, and they are implemented using 
Mujoco physics engine \citep{todorow_mujoco_2012}. We use \(r_i\) to denote the ith reward 
component. When computing Hypervolume, we use the vector of zeros \(\bm{0}_d\) as our reference 
point in all environments.

\textbf{Ant-v4} has an observation space \(\mathcal{S} \in \R^{27}\), and action-space \(\mathcal{A} \in \R^8\). The agent is limited to 500 timesteps
in this environment. The environment has two objectives, x-axis speed and y-axis speed: 
\begin{align*}
   r_1 &= v_x + C\\
   r_2 &= v_y + C
\end{align*}
where \(v_x\) is the speed in \(x\) direction, \(v_y\) is the speed in \(y\) direction and \(C = 1 - 0.5 \sum_i a^2_i\) is
combination of alive bonus and energy efficiency, defined as the squared sum of actions for each actuator.

\textbf{HalfCheetah-v4} has an observation-space \(\mathcal{S} \in \R^{17}\) and action-space \(\mathcal{A} \in \R^6\). The agent is limited to 500 timesteps
in this environment. The environment has two objectives, forward speed and energy efficiency:
\begin{align*}
   r_1 &= \min(v_x, 4) +  C \\ \\
   r_2 &= 4 -\sum_{i} a_i^2 +C
\end{align*}
where \(v_x\) is the speed in \(x\) direction, \(C = 1\) is alive bonus and \(a_i\) is the action of each actuator.

\textbf{Hopper-v4} has an observation-space \(\mathcal{S} \in \R^{11}\) and action-space \(\mathcal{A} \in \R^3\). The agent is limited to 500  timesteps
in this environment. In the \emph{two objective} cases, the rewards are forward speed and jumping height:
\begin{align*}
   r_1 &= 1.5v_x +  C \\
   r_2 &= 12 (h - h_{\mathrm{init}}) + C
\end{align*}
where \(v_x\) is the speed in \(x\) direction, \(C = 1 - 0.0002\sum_i a_i^2\) combines alive bonus and 
energy efficiency, while \(h\) is the current height and \(h_{\mathrm{init}}\) is the initial height.

In the \emph{three objective} version, the rewards are forward speed, jumping height and energy efficiency
\begin{align*}
   r_1 &= 1.5v_x +  C \\
   r_2 &= 12 (h - h_{\mathrm{init}}) + C \\
   r_3 &= 4 - \sum_i a_i^2 + C
\end{align*}
where \(C = 1\) is the alive bonus, and the rest of the symbols have 
same meaning as in the two-objective configuration.

\textbf{Swimmer-v4} has an observation-space \(\mathcal{S} \in \R^{8}\) and action-space \(\mathcal{A} \in \R^2\). The agent is limited to 500 timesteps
in this environment. The environment has two objectives, forward speed and energy efficiency:
\begin{align*}
   r_1 &= v_x \\
   r_2 &= 0.3 - 0.15\sum_{i} a_i^2
\end{align*}
where \(v_x\) is the speed in \(x\) direction and \(a_i\) 
is the action of each actuator.

\textbf{Walker2d-v4} has an observation-space \(\mathcal{S} \in \R^{17}\) and action-space \(\mathcal{A} \in \R^6\). The agent is limited to 500 timesteps
in this environment. The environment has two objectives, forward speed and energy efficiency:
\begin{align*}
   r_1 &= v_x + C\\
   r_2 &= 4 - \sum_{i} a_i^2 + C
\end{align*}
where \(C = 1\) is the alive bonus, \(v_x\) is the speed in \(x\) direction 
and \(a_i\) is the action of each actuator.

\textbf{Humanoid-v4} has an observation-space \(\mathcal{S} \in \R^{376}\) and action-space \(\mathcal{A} \in \R^{17}\). The agent is limited to 500 timesteps
in this environment. The environment has two objectives, forward speed and energy efficiency:
\begin{align*}
   r_1 &= 1.25v_x + C\\
   r_2 &= 3 - 4\sum_{i} a_i^2 + C
\end{align*}
where \(C = 3\) is the alive bonus, \(v_x\) is the speed in \(x\) direction and \(a_i\) is the action of each actuator.

\section{Further Empirical Evaluations}\label{appendix:pareto_fronts}
\begin{figure}[h]
    \centering
    \includegraphics[alt={Hypervolume (top) anb EUM (bottom) as function of timesteps in 4 environments. Momba beats the runner-up, PGMORL in 3 cases, while matching it in the challenging Swimmer environment. In all cases, Momba requires a fraction of timesteps compared to PGMORL.},width=\linewidth]{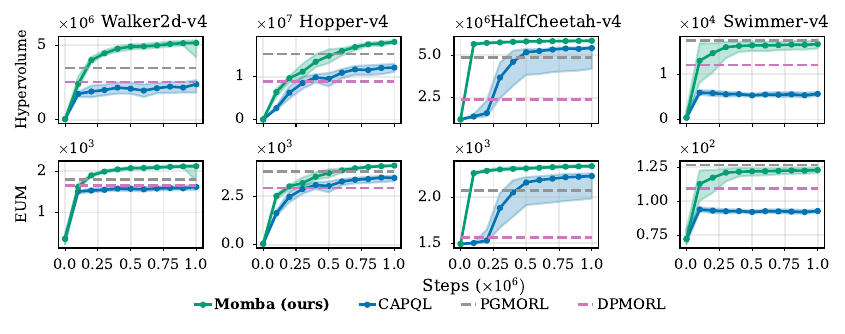}
    \caption{
        HV (top) and EUM (bottom) (IQM and 95\% SBCIs over 10  seeds) during training.
        Horizontal lines indicate the final performance of PGMORL and DPMORL.
    }
    
    \label{fig:app_sample_efficiency}
\end{figure}

\begin{figure}[h]
    \centering
    \includegraphics[alt={Hypervolume of Momba as function of timesteps in all environments with UTD ratios 1, 2, 4 and 8. The improvements in convergence speed are modest in all cases.},width=\linewidth]{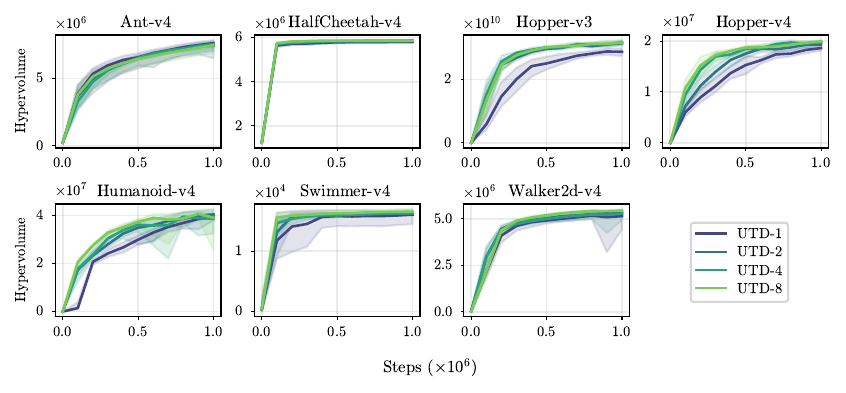}
    \caption{
        HV of Momba during training on all environments 
        (IQM and 95\% SBCIs over 10 seeds) with various UTD-ratios. 
    }
    \label{fig:app_utd_scaling}
\end{figure}

\begin{figure}[h]
    \centering
    \includegraphics[alt={Solution sets for Humanoid (left), Swimmer (center) and Hopper-v3 (right). In Humanoid and Swimmer, Momba produces best solution set, followed by PGMORL. In Hopper-v3, PGMORL has better coverage over the solution space.},width=\linewidth]{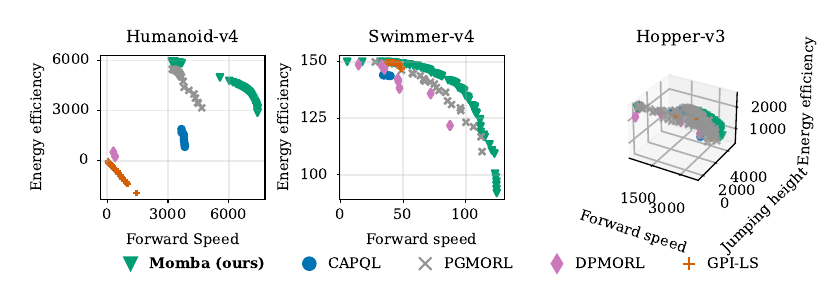}
    \caption{
        Generated solution sets in Humanoid (left), Swimmer (center), and Hopper-v3 (right).
        Humanoid and Swimmer remain difficult environments for most of the algorithms. 
    }
    \label{fig:app_solution_sets}
\end{figure}

In this section, we showcase the training curves (\Figref{fig:app_sample_efficiency}), 
UTD-scaling results (\Figref{fig:app_utd_scaling}) and obtained solution sets (\Figref{fig:app_solution_sets})
in the environments not shown in the main text. 

\Figref{fig:app_sample_efficiency} showcases that Momba outperforms its base algorithm, CAPQL, 
in all environments. We also outperform PGMORL, while requiring a fraction of the training steps
in all environments except Swimmer, \camera{%
    where Momba matches the performance of PGMORL. Interestingly, CAPQL converges very quickly
    to a sub-optimal policy in Swimmer, again highlighting the effectiveness of the proposed approach.%
}

\Figref{fig:app_utd_scaling} displays the UTD-scaling in different environments. 
While we see that the Hopper and Humanoid environments benefit from higher UTD, the performance improvements 
in other environments are limited. \camera{We note that in HalfCheetah, most algorithms
start to approach the theoretical upper bound of \(6.25 \times 10^6\) for hypervolume (see \Figref{fig:app_sample_efficiency}), indicating that the environment may be too easy for the current methods.}

Lastly, \Figref{fig:app_solution_sets} displays the generated solution sets in Humanoid, Swimmer, and Hopper-v3.
Humanoid remains a difficult environment, as none of the algorithms can obtain a good coverage of the solution
set. \camera{%
    On the other hand, both CAPQL and GPI-LS, representing the general policy methods, struggle to find 
    policies that can move the actor forward, instead collapsing and generating a set of policies that all optimize for the energy efficiency.
    On the contrary, Momba recovers a solution set that is on par with PGMORL, being the only general policy method to do so.%
}

%
%

\end{document}